%% file: main.tex
\documentclass{article}

\usepackage[final,main]{neurips_2025}
\usepackage[T1]{fontenc}
\usepackage[utf8]{inputenc}
\usepackage{graphicx}
\usepackage{booktabs}
\usepackage{amsmath}
\usepackage{amssymb}
\usepackage{array}
\usepackage{tabularx}
\usepackage{multirow}
\usepackage{longtable}
\usepackage{listings}
\usepackage{xcolor}
\usepackage{hyperref}

\newcolumntype{L}{>{\raggedright\arraybackslash}X}
\newcolumntype{P}[1]{>{\raggedright\arraybackslash}p{#1}}

\lstdefinestyle{codestyle}{
  basicstyle=\ttfamily\footnotesize,
  breaklines=true,
  frame=single,
  columns=fullflexible,
  keepspaces=true,
  showstringspaces=false
}

\title{Oracle Agent Memory as an Enterprise Memory Substrate for Long-Horizon AI Agents}

\author{%
  \normalfont
  Richmond Alake, Cesare Bernardis, Paul Cayet, Luca Engel, \\
  Damien Hilloulin, Sungpack Hong, Allen Hosler, Nickolas Kavantzas, \\
  Ingo Kossyk, Son Le, Rhicheek Patra, Kartik Talamadupula, Valentin Venzin \\[0.75em]
  \textbf{Oracle}
}

\makeatletter
\renewcommand{\@noticestring}{}
\makeatother

\begin{document}

\maketitle

\input{sections/00_abstract}
\input{sections/01_introduction}
\input{sections/02_memory_taxonomy}
\input{sections/03_related_work}
\input{sections/04_architecture}
\input{sections/05_integration_workflow}
\input{sections/06_use_cases}
\input{sections/07_evaluation}
\input{sections/08_discussion}
\input{sections/09_limitations}
\input{sections/09_future_work}
\input{sections/09_conclusion}

\bibliographystyle{plain}
\bibliography{references}

\clearpage
\appendix

\input{sections/10_appendix_setup}
\input{sections/11_appendix_lifecycle}
\input{sections/12_appendix_search}

\end{document}

%% file: sections/00_abstract.tex
\begin{abstract}
Agent memory is a systems problem for long-horizon agents. Practical deployments require retention of task state across extended conversations, recovery of user-specific facts and preferences across sessions, and accumulation of procedural knowledge from prior outcomes. These requirements extend beyond document retrieval: a memory layer must determine which interactions become durable state, how that state is scoped, how it is retrieved under latency constraints, and how it is revised or removed over time. This report studies Oracle Agent Memory as a database-native memory substrate built on Oracle Database. Three themes organize the discussion: memory as a lifecycle spanning ingestion, extraction, consolidation, retrieval, summarization, and revision or removal; a layered architecture that separates an active memory core from a passive memory-store interface with explicit scope control across users, agents, and threads; and evaluation methodology in which downstream task accuracy is complemented by memory-centric measures such as evidence retrieval, recall, latency, and estimated token use. The report summarizes LongMemEval results (reaching $93.8$\% accuracy), compares Oracle Agent Memory against flat-history baselines (using about $10.7\times$ fewer tokens) and published or reported external baselines where available, and closes with implementation-oriented appendix material covering setup, thread lifecycle, and search semantics.
\end{abstract}

%% file: sections/01_introduction.tex
\section{Introduction}
Agent memory has become a central systems concern for enterprise agents because most useful deployments extend beyond a single bounded interaction. Agents accumulate user preferences, factual context, intermediate plans, and traces of prior execution. A support agent may need incident history and account-specific preferences across weeks. A coding agent may need prior design decisions, failed experiments, and reusable fixes across sessions. An analytics or NL2SQL agent may need domain rules, recurring join patterns, and clarifications of ambiguous business terminology. The core problem is therefore not mere transcript recall, but maintenance of heterogeneous memory under explicit scope, retention, and retrieval constraints.

This requirement alters the design space. A memory subsystem must synchronize conversation history and tool traces, extract durable facts or guidelines from raw interaction, materialize compact short-term context for subsequent turns, retrieve relevant prior evidence under latency constraints, and enforce scope boundaries across users, agents, and conversations. The subsystem must also accommodate revision over time as facts become stale, preferences change, and experiential memory is updated from new outcomes. Memory is therefore more naturally modeled as a managed lifecycle than as a single retrieval interface.

Recent survey work supports this view. Memory in agentic systems spans multiple forms, functions, and temporal dynamics, and is not adequately characterized by a simple short-term versus long-term dichotomy \cite{memorysurvey2025,memorymechanismsurvey2024}. At the same time, implementations remain fragmented across virtual-context managers, reflection buffers, memory-specific services, vector stores, graph layers, and framework abstractions \cite{memgpt2024,reflexion2023,mem0,zep,amem2025,letta,langmem}. In enterprise settings, such fragmentation complicates governance, security, data locality, reliability, and integration with the application logic that governs the agent loop.

Oracle Agent Memory is motivated by the observation that these memory requirements align closely with database requirements. Messages, summaries, facts, profiles, and related memory artifacts require persistence, indexing, scope-aware retrieval, and compatibility with structured and semi-structured enterprise data. Oracle Database already provides relational storage, JSON, vectors, transactions, and operational controls. The system studied here consequently treats the database as the substrate for an explicit memory layer rather than as an external sink for embeddings.

The remainder of the report is organized as follows. Section~2 formulates memory as a lifecycle and introduces the taxonomy used throughout the paper. Section~3 situates Oracle Agent Memory within prior memory systems and benchmarks. Sections~4 through~6 describe the architecture, integration workflow, and representative workloads. Sections~7 and~8 examine evaluation methodology, including the current LongMemEval-based results and the broader question of how a memory subsystem should be measured. The report then states current limitations, identifies future work, and concludes.

%% file: sections/02_memory_taxonomy.tex
\section{Memory Taxonomy and Problem Framing}
We organize the memory problem around three operational categories that map well onto the broader survey literature and prior agent-memory systems \cite{memorysurvey2025,memorymechanismsurvey2024,generativeagents2023,reflexion2023}. First, \emph{short-term or working memory} captures the active state of a task: thread summaries, context cards, unresolved subgoals, recent actions, and the minimal contextual state needed to keep an agent coherent over long interactions. Second, \emph{long-term factual memory} stores stable information such as user preferences, known facts, rules, profile-like attributes, and domain knowledge that should remain available across sessions. Third, \emph{long-term experiential or procedural memory} captures outcomes, strategies, lessons, and guidelines learned from prior execution traces or feedback.

These categories differ materially at runtime. Working memory is highly dynamic and tightly coupled to the current interaction. It must be refreshed frequently, remain compact enough for prompt injection, and preserve the information that governs the next turn rather than the full conversational trace. In Oracle Agent Memory, this role is represented primarily by thread summaries and context cards.

Factual memory has a different operational profile. Facts and preferences may originate in a particular interaction, but once extracted they frequently need to persist beyond the originating thread. A statement such as ``the user prefers concise answers'' or ``this analyst uses fiscal quarters aligned to a custom calendar'' becomes durable context that must remain searchable, attributable to the correct user or agent, and compatible with retrieval and governance policies. Factual memory therefore emphasizes persistence, indexing, and scope-aware lookup rather than turn-level refresh.

Procedural or experiential memory poses a more difficult inference problem. Storage of an event alone is insufficient; the system must identify which lesson, strategy, or guideline should be retained. In an NL2SQL workflow the durable artifact may be a rule for interpreting ``average spending'' or a learned preference for excluding zero-spend customers unless explicitly requested otherwise. In a troubleshooting workflow it may be evidence that a remediation path resolved a recurring incident class. Procedural memory consequently depends on extraction, consolidation, and, in some cases, reflection over outcomes rather than direct storage of raw messages alone.

A lifecycle formulation is more informative than a static type hierarchy. In a typical flow, new messages or tool traces enter the system as raw records. Those records may trigger extraction of candidate memories. The system may summarize the current thread for short-term use, consolidate recurring evidence into longer-lived memories, index text-bearing records for semantic retrieval, and later retrieve or recompose a relevant subset for a new task. Over time, memories may need to be edited, merged, de-scoped, or removed as evidence changes.

This lifecycle exposes several systems problems. The first is \emph{selection}: not every message should become durable memory. Over-aggressive extraction increases clutter, contradiction, and retrieval cost, whereas under-extraction yields cold starts and repeated clarification. The second is \emph{representation}: some memory is naturally stored as narrative text, some as profile records, some as message history, and some as more structured or graph-like artifacts. The third is \emph{scope}: a memory may apply only within one thread, across many threads for one user, across an agent serving many users, or globally. The fourth is \emph{evolution}: memory may become stale, conflict with newer evidence, or encode undesirable feedback in the absence of editing and forgetting mechanisms.

Enterprise settings sharpen these constraints. Memory systems require governed access to business data, predictable behavior across user and agent boundaries, and unified control over structured, semi-structured, and vectorized representations. They must also preserve the distinction between persistence and inference. Raw thread messages remain evidentiary records even when summaries or extracted memories are later revised. Oracle Agent Memory addresses these requirements by combining thread-level workflows for active memory management with a broader store abstraction capable of holding messages, memories, profiles, and future artifacts under a common scope model.

The subsequent sections use this lifecycle perspective throughout. The architecture section examines how the system supports each stage of memory management. The workflow section studies how thread operations, search, and explicit memory insertion map onto an agent harness. The evaluation section distinguishes downstream answer quality from the quality of selection, persistence, and retrieval within the memory lifecycle itself.

%% file: sections/03_related_work.tex
\section{Related Work and Benchmarks}
\subsection{Memory Mechanisms in Language Agents}
Recent surveys characterize memory in language-agent systems as a family of mechanisms rather than a single persistence primitive \cite{memorysurvey2025,memorymechanismsurvey2024}. The relevant design space includes short-term state used to maintain local coherence, long-term semantic memory for facts and preferences, episodic memory for prior interactions, and procedural memory derived from previous outcomes. This framing is consistent with agent systems that explicitly maintain observations, plans, reflections, or retrieved evidence outside the immediate model context \cite{generativeagents2023,reflexion2023,memgpt2024}.

Several influential systems established memory as part of the agent loop. Generative Agents maintains a memory stream over observations, retrieves memories by relevance, recency, and importance, and uses reflection to synthesize higher-level inferences from accumulated experience \cite{generativeagents2023}. Reflexion stores verbal feedback in an episodic memory buffer and reuses it to improve subsequent trials without updating model weights \cite{reflexion2023}. ReAct and Toolformer are not memory systems in the narrow sense, but they clarify the broader agent setting in which memory must coexist with tool use, external actions, and iterative control flow \cite{react2023,toolformer2023}. These systems motivate a memory substrate that stores more than raw dialogue: traces, outcomes, summaries, and tool-mediated state can all become candidates for later retrieval.

\subsection{Retrieval-Augmented and Non-Parametric Memory}
Agent memory also builds on retrieval-augmented language modeling. RAG combines a parametric sequence-to-sequence model with a dense non-parametric memory of retrieved passages, making retrieval a first-class component of generation \cite{rag2020}. Dense Passage Retrieval showed that learned dense embeddings can replace sparse lexical matching for open-domain question answering at scale \cite{dpr2020}. REALM integrates retrieval into language-model pretraining, while RETRO conditions language models on retrieved chunks from very large corpora \cite{realm2020,retro2022}. These papers establish the core mechanism used by many memory systems: encode text-bearing records, search a vector index, and condition generation on retrieved evidence.

Conversational memory differs from document RAG in several respects. The stored artifacts are often inferred rather than copied from a canonical document. They are scoped to a user, thread, agent, tenant, or workflow rather than globally valid. They may be revised as preferences change or as new evidence supersedes old state. They also include derived artifacts such as summaries, context cards, and reflections. A generic retrieval pipeline can approximate these behaviors, but first-class notions of scope, lifecycle, and record type reduce the amount of application-specific policy that must be rebuilt around the retriever.

\subsection{Long Context and Virtual Context Management}
Long-context modeling is complementary to external memory, but it does not eliminate the need for memory management. Architectures such as Longformer reduce the cost of processing long documents through sparse attention patterns \cite{longformer2020}. However, empirical work on long-context use shows that models can fail to use information reliably when relevant evidence appears in the middle of long inputs \cite{lostmiddle2024}. This limitation is important for agent memory because a flat transcript may contain the necessary fact while still presenting it in a form that is difficult for the model to exploit.

MemGPT addresses the same constraint through virtual context management: the model operates over a bounded active context while moving information between working context and longer-lived storage tiers \cite{memgpt2024}. This operating-system analogy is directly relevant to enterprise agent memory. The objective is not simply to maximize the amount of text sent to the model; it is to decide which state should be active, which state should be stored durably, and which records should be retrieved for a given task.

\subsection{Contemporary Agent-Memory Systems}
Recent memory systems make these ideas more explicit. Mem0 extracts, consolidates, and retrieves salient conversational information, including graph-based variants for relational memory \cite{mem0}. Zep represents agent memory through a temporal knowledge graph that supports evolving entities and relationships over time \cite{zep}. A-Mem proposes agentic memory organization in which new memories can trigger linking and evolution of prior records \cite{amem2025}. Letta and LangMem represent the implementation ecosystem around these ideas: memory abstractions are increasingly exposed as reusable components for agent builders rather than as ad hoc prompt-management code \cite{letta,langmem}.

These systems differ in emphasis. Some prioritize personalization, some graph-structured temporal recall, some explicit context paging, and some framework integration. They nevertheless share several design pressures: extraction must be selective, retrieval must be scoped, summaries must be compact, and memory updates must remain controllable. Oracle Agent Memory is situated within this systems gap. It treats the database, the memory-store contract, and the active memory-management layer as a unified substrate rather than as separate services assembled around an agent framework.

\begin{table*}[t]
\centering
\small
\begin{tabularx}{\textwidth}{P{0.19\textwidth} P{0.22\textwidth} L}
\toprule
Benchmark & Primary emphasis & What it tests \\
\midrule
LoCoMo \cite{locomo} & Very long conversational memory & Long-horizon question answering, event summarization, temporal and causal reasoning, and multimodal dialogue over conversations spanning many sessions. \\
LongMemEval \cite{longmemeval} & Multi-session chat-assistant memory & Information extraction, multi-session reasoning, temporal reasoning, knowledge updates, and abstention over scalable timestamped chat histories. \\
Deep-memory retrieval \cite{memgpt2024,zep} & Prior-session evidence retrieval & Retrieval of information answerable only from knowledge acquired in earlier conversation sessions. \\
MemoryAgentBench \cite{memoryagentbench} & Incremental agent interaction & Accurate retrieval, test-time learning, long-range understanding, and conflict resolution across multi-turn agent trajectories. \\
\bottomrule
\end{tabularx}
\caption{Representative benchmarks and evaluation settings for long-horizon agent memory.}
\label{tab:memory-benchmarks}
\end{table*}

\subsection{Memory Benchmarks}
These benchmarks test different views of memory quality. LoCoMo evaluates very long-term conversational understanding through question answering, event summarization, and dialogue generation grounded in extended multi-session histories \cite{locomo}. LongMemEval targets the retrieval regime faced by memory-enabled chat assistants: evidence is distributed across timestamped sessions, relevant facts are sparse, and questions may require cross-session synthesis, temporal reasoning, knowledge updates, or abstention \cite{longmemeval}. MemoryAgentBench moves further toward online agent behavior by testing incremental multi-turn interactions in which memory must support repeated decision making rather than a single isolated answer \cite{memoryagentbench}.

No single benchmark is sufficient. Long-context comprehension, retrieval recall, answer correctness, update handling, abstention, and cost behavior stress different subsystems. A system can improve downstream question answering through broad retrieval fan-out while increasing latency and token cost. Conversely, aggressive summarization can reduce cost while degrading recoverability of rare evidence. For this reason, the evaluation in Section~7 combines LongMemEval accuracy with token-growth measurements, flat-history comparisons, threshold sweeps, and prompt-caching analysis.

%% file: sections/04_architecture.tex
\section{Oracle Agent Memory Architecture}
Oracle Agent Memory is organized as a layered stack rather than a single API surface. The design separates an \emph{active memory core}, responsible for thread synchronization, extraction, summarization, and search orchestration, from a \emph{memory store} interface that supports passive persistence, scoped retrieval, and compatibility with third-party memory providers. This division reflects two requirements common in enterprise memory systems: an opinionated path for recurrent agent workflows and a stable storage contract that is not tied to one memory algorithm.

\begin{figure*}[t]
  \centering
  \includegraphics[width=0.9\textwidth]{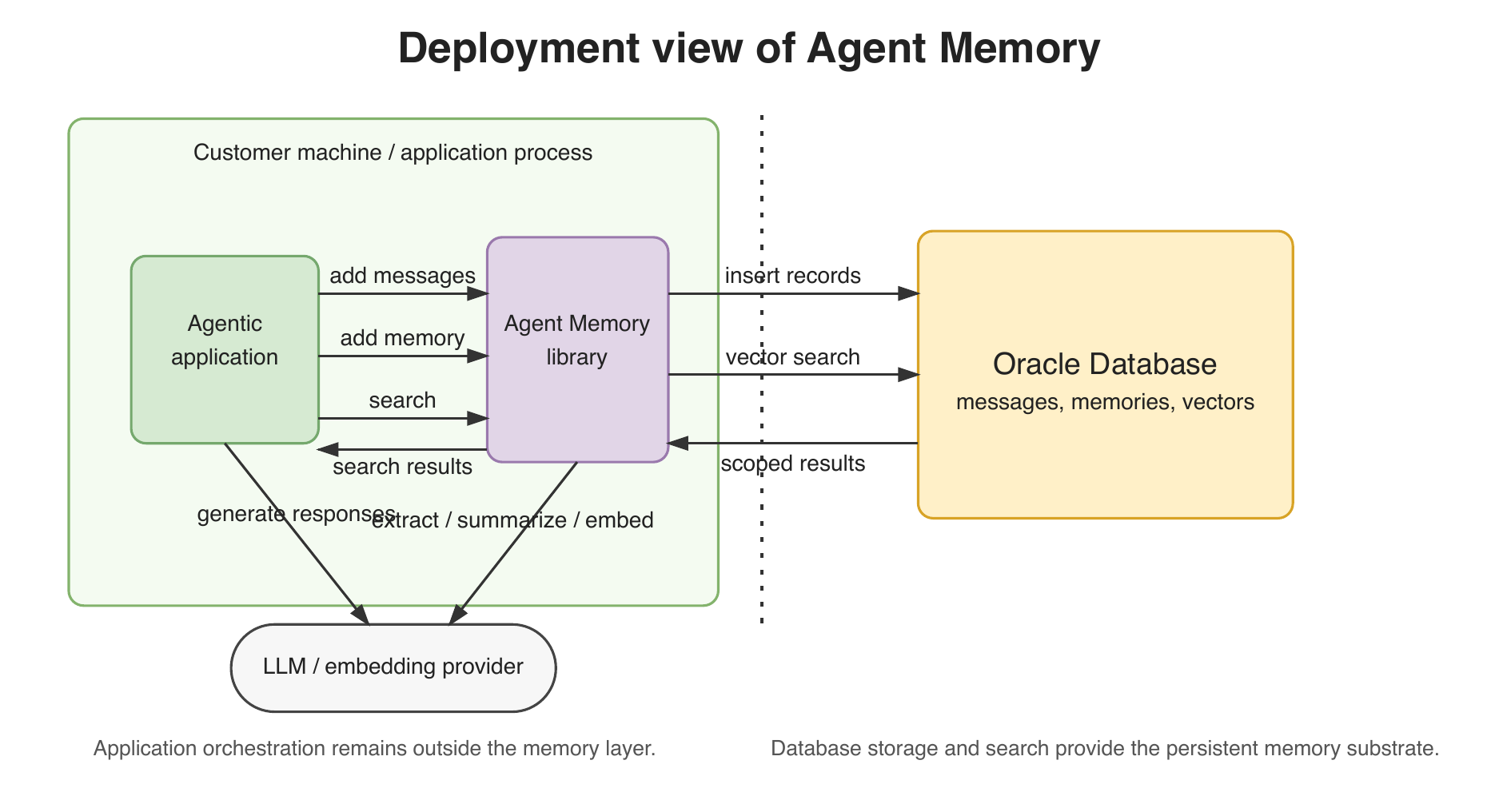}
  \caption{Deployment-oriented view of Oracle Agent Memory. The customer application interacts with the Oracle Agent Memory library, which in turn coordinates storage and search against Oracle Database while calling external or adjacent LLM and embedding services for extraction, summarization, and indexing.}
  \label{fig:architecture-deployment}
\end{figure*}

Figure~\ref{fig:architecture-deployment} shows the deployment split between the customer application process, the Oracle Agent Memory library, Oracle Database, and external model endpoints. The figure identifies the boundary between application orchestration, memory policy, and database-backed persistence. Message insertion, memory insertion, and search pass through the memory layer, whereas LLMs and embedding models are invoked only for extraction, summarization, and indexing.

This deployment model has several consequences. The application or agent harness remains responsible for end-user interaction, tool routing, and higher-level workflow logic. Oracle Agent Memory serves as the subsystem that persists thread state, materializes short-term context, extracts or accepts durable memories, and retrieves scoped evidence. Oracle Database is the persistence and search substrate for the memory layer rather than an archival endpoint appended to the agent stack.

A DB connection or pool is required for persistence and retrieval. An embedding model endpoint is required for semantic indexing and search. An LLM endpoint is optional at initialization time but becomes necessary when the system performs memory extraction, summarization, or LLM-assisted post-processing. This separation allows one API surface to support both an active configuration, in which thread updates trigger extraction and summary refresh, and a more conservative configuration, in which the application persists messages and memories explicitly without automated inference.

\begin{figure*}[t]
  \centering
  \includegraphics[width=\textwidth]{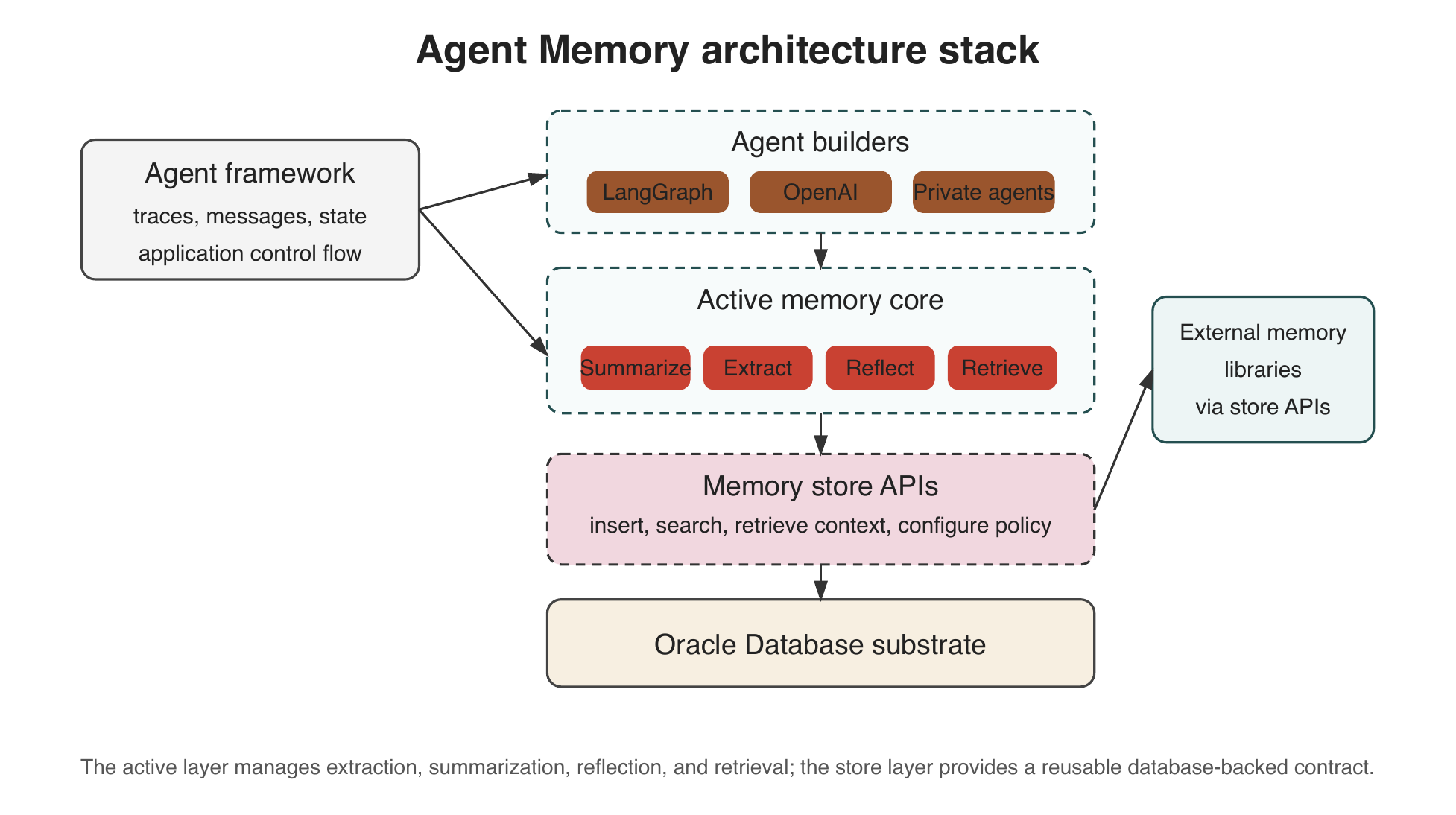}
  \caption{End-to-end memory stack for long-horizon memory-enabled agents. The stack distinguishes agent builders, memory-core logic, memory-store APIs, and the database substrate, while also showing the compatibility path for external memory libraries.}
  \label{fig:end-to-end-stack}
\end{figure*}

The broader architectural stack is captured in Figure~\ref{fig:end-to-end-stack}. Agent frameworks collect traces, messages, and state from execution. The memory core performs summarization, extraction, and retrieval orchestration. The memory-store APIs expose insert, search, and retrieval operations. Oracle Database provides the converged substrate---relational tables, JSON, vector indexes, transactional metadata, and schema patterns that can support future graph-aware memory structures---that keeps memory close to the underlying enterprise data. This layered decomposition accommodates both Oracle-specific memory algorithms and storage compatibility for third-party memory libraries.

The active layer is centered on the \texttt{OracleAgentMemory} client and its thread-oriented workflow. It provides interfaces for creating or reopening threads, adding messages, building context cards, generating summaries, adding explicit memories, managing actor profiles, and executing scoped search. Message insertion can trigger memory extraction. Summarization APIs provide compact short-term context. Search is available both from the thread handle and at the global client level. The resulting API model aligns directly with the structure of agent turns.

The passive layer is the \emph{memory store}. It persists records, exposes scoped retrieval, and performs semantic search over the stored corpus. The store contract is intended to support Oracle Agent Memory itself as well as third-party memory providers that use Oracle Database as a storage and search backend. It is termed ``passive'' because it accepts records and returns records, but does not itself decide what should be extracted or when summaries should be recomputed. This distinction allows the storage substrate to remain reusable even when active memory-management policy changes.

\begin{figure*}[t]
  \centering
  \includegraphics[width=0.92\textwidth]{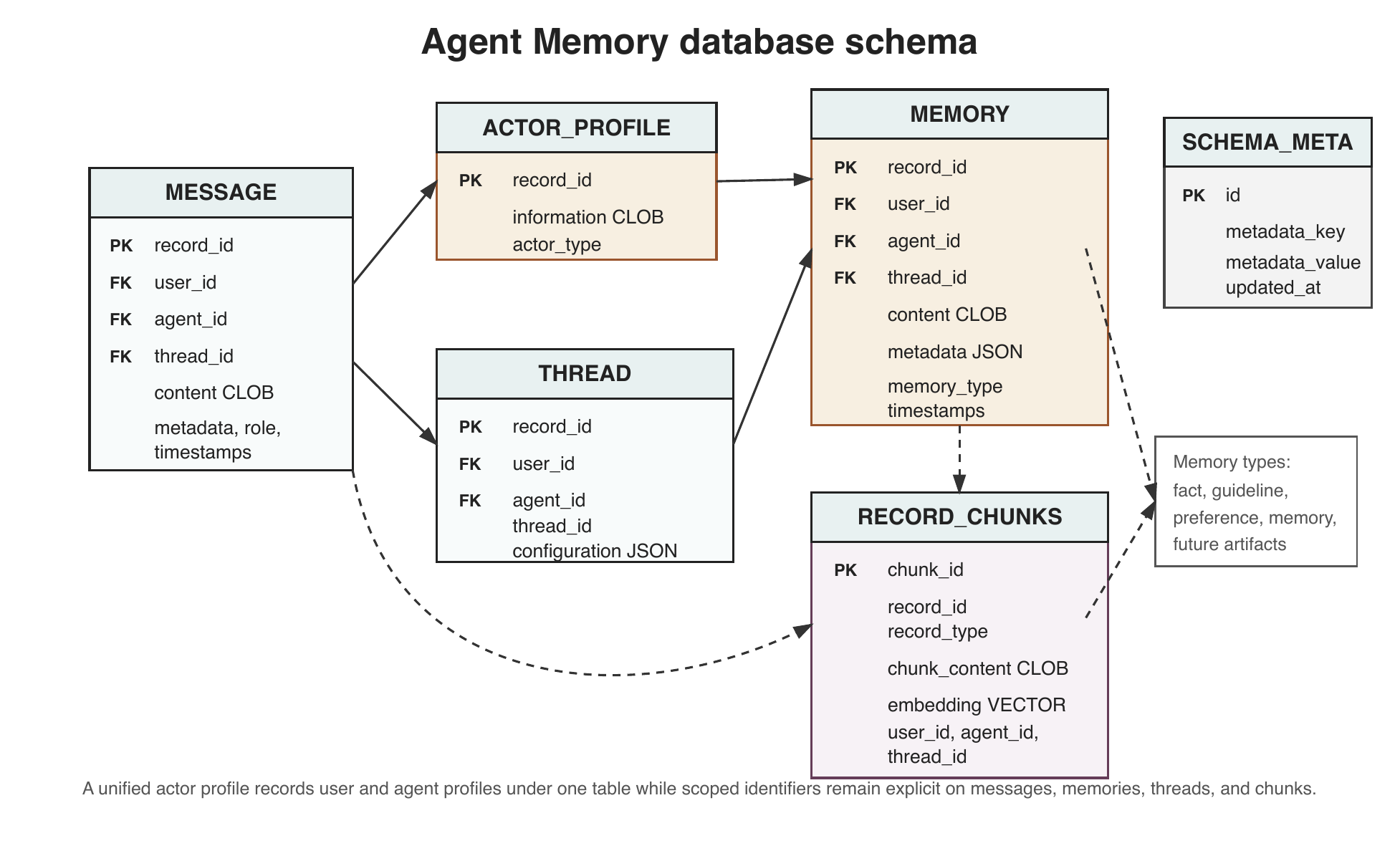}
  \caption{Schema-level view of the Oracle Agent Memory data model. Messages, threads, actor profiles, memories, and chunked vectorized records share a common relational backbone with explicit user, agent, and thread scope.}
  \label{fig:schema-data-model}
\end{figure*}

Figure~\ref{fig:schema-data-model} makes the storage model concrete. The design centers on a small set of record-bearing entities: \texttt{THREAD}, \texttt{MESSAGE}, \texttt{MEMORY}, \texttt{ACTOR\_PROFILE}, and chunked vectorized records for semantic retrieval. The \texttt{ACTOR\_PROFILE} table represents profile records for users and agents under a single relation, with \texttt{actor\_type} distinguishing the actor class. The schema reinforces several architectural priorities:
\begin{itemize}
  \item scoped storage across user, agent, and thread boundaries;
  \item a unified relational-plus-vector view over raw messages and distilled memories;
  \item a path for multiple memory types, including facts, guidelines, preferences, and future artifact types;
  \item explicit chunk storage to support semantic search over long text-bearing records.
\end{itemize}

The record model normalizes several classes of memory artifact under one storage contract. Messages are stored as raw records. Durable memories are stored as separate records. Actor profiles can be inserted explicitly and associated with user or agent scope. Index text and embeddings support retrieval, whereas timestamps, roles, and metadata preserve contextual detail. The same retrieval path can therefore search both conversational evidence and derived memories, or restrict itself to one record type when stricter control is required.

The ingestion and retrieval paths may be more complex than a single vector lookup. On ingestion, a new event can trigger retrieval of related entities, optional graph traversal, asynchronous updates to existing entities or relations, and chunking of raw text into separate search records. On retrieval, semantic search can be combined with structured filters, record-type selection, and optional post-processing. Although the MVP emphasizes conversational memory and vector search, the architecture leaves room for richer search and evolution pipelines.

Scoping is the second architectural pillar. The system treats user, agent, and thread identifiers as first-class storage and retrieval dimensions. A memory may be tied to a specific thread, shared across threads for one user, or associated with an agent across many users. Search strictness is controlled by exact-match flags on each scope dimension, allowing the caller to trade precision for recall in a predictable manner. A support assistant may require strict thread matching while summarizing the current incident, yet broader user-level recall when retrieving long-term preferences or repeated issue patterns.

Database-native memory also changes the enforcement point for security policy. Agent memory can contain private conversation turns, inferred preferences, procedural lessons, and retrieved enterprise facts. If these artifacts are stored in an application-side vector service with a separate authorization model, the system must replicate identity, role, and policy context outside the primary data platform. Oracle Deep Data Security instead emphasizes identity-aware access control in the database layer, including policy enforcement over SQL access paths and fine-grained controls over rows, columns, and cells \cite{oracledeepdatasecurityfeature,oracledeepdatasecurityblog2026,oracledeepdatasecuritybrief2026}. For memory retrieval, this means that semantic search and structured filtering can be composed with database policy rather than treated as a post-hoc application filter. For example, a support agent answering a request for \texttt{user\_123} in \texttt{thread\_incident\_17} may search user-scoped memories, agent-scoped guidelines, and thread-local messages. Database policy should first exclude rows that the acting identity is not allowed to access; memory scope filters should then restrict the eligible rows to the requested user, agent, or thread. A denied record should be absent from the result set because it is not admissible for the identity and task context, not merely because it was filtered after retrieval.

Scope identifiers and exact-match flags are retrieval controls and should not be treated as a complete authorization boundary by themselves. Production deployments should enforce authorization through database privileges, database policy, application identity propagation, and audit controls, with memory scope filters serving as part of the retrieval contract.

This property is particularly important for multi-user and multi-agent memory. A retrieval call may combine user-scoped facts, agent-scoped guidelines, thread-local evidence, and adjacent business records. The intended result is not simply the nearest vectors, but the nearest admissible records for the acting identity and task context. Placing memory in Oracle Database permits the memory layer to inherit database governance mechanisms, auditability, and policy locality while still using vectors, JSON, relational predicates, and future graph-oriented relationships as retrieval primitives. The architecture therefore treats security and retrieval as coupled systems concerns rather than independent middleware responsibilities.

Short-term memory is handled through context cards and summaries rather than through a monolithic conversation-memory object. Context cards capture the thread topic, key current facts, and other information relevant to the next agent turn. Thread summaries compress the conversation and can be bounded or truncated according to token budget. The separation between these two artifacts reflects their different roles: summaries provide compression, whereas context cards provide prompt-ready state for the current thread.

Finally, schema creation and compatibility are expressed as explicit policy rather than hidden side effects. The design includes several schema policies, including a default mode that validates an existing managed schema without executing DDL, as well as more permissive modes that create or recreate managed objects when appropriate. In enterprise deployments, memory infrastructure must coexist with operational controls, least-privilege grants, and compatibility checks such as schema-version or embedding-dimension validation. Oracle Agent Memory therefore encodes these operational constraints directly in the architecture.

%% file: sections/05_integration_workflow.tex
\section{Integration and Developer Workflow}
The thread is the primary integration primitive for agent memory. Conversation threads and messages are synchronized into memory, where automatic extraction can promote durable memories from message history. Explicit \texttt{add\_memory()} and \texttt{search()} tools may also be exposed to the agent, so that memory updates and retrieval participate directly in the tool-use loop rather than only in a fixed middleware pipeline. The resulting workflow combines automated synchronization with explicit agent- or application-driven memory operations.

At a high level, the workflow has four recurring stages:
\begin{enumerate}
  \item initialize the memory client with a database connection, embedder, and optional LLM;
  \item create or reopen a thread with optional user and agent scope;
  \item add messages so the system can persist history and run extraction/summarization flows;
  \item retrieve context via scoped search, context cards, or thread summaries before the next agent turn.
\end{enumerate}

The first stage is setup. The application instantiates \texttt{OracleAgentMemory} with a DB connection or pool, an embedder, and optionally an LLM. This configuration determines whether memory extraction is active or disabled, whether managed schema objects may be created automatically, and which model backends are used for embeddings and generation. These choices determine whether the system operates as an active memory layer with automatic extraction and summarization or as a more explicit persistence and retrieval service.

The second stage is thread lifecycle management. The application creates a thread, usually attaching optional \texttt{user\_id} and \texttt{agent\_id} scope. The thread handle then becomes the stable unit for turn-by-turn work. It can be reopened later, deleted when necessary, and used for message insertion, summary generation, context-card generation, explicit memory writes, and scoped search. Since most agent interactions are naturally organized around conversations or task threads, a thread-centric interface aligns with existing orchestration logic.

The third stage is message synchronization and memory construction. As new user or assistant messages arrive, the harness appends them through \texttt{add\_messages(...)}. These messages are stored as raw records and remain searchable later. When extraction is enabled, message insertion may also trigger short-term summary refresh and durable memory extraction. At this stage, raw interaction is transformed into reusable state.

The fourth stage is retrieval and prompt conditioning. Before an agent turn, the harness may call \texttt{get\_context\_card()} to obtain a compact structured representation of thread state, or \texttt{get\_summary(...)} to compress a longer history under a token budget. During tool use, the harness or the agent may call \texttt{search(...)} to retrieve semantically relevant messages and memories. Context cards and summaries provide proactive short-term memory; search provides reactive long-term recall.

This design combines automatic and explicit memory operations. Automatic extraction allows the system to build factual memory and summaries from ordinary conversation flow. Explicit APIs allow the application or agent to persist durable artifacts such as user preferences, outcomes, or plans. Three integration patterns follow naturally. In a middleware-centric pattern, the harness synchronizes every thread and injects summaries or context cards before each turn. In a tool-centric pattern, the agent decides when to invoke memory search or add-memory tools. In a hybrid pattern, the harness manages thread synchronization and short-term context automatically, whereas the agent uses explicit memory tools for selected writes and targeted recall. These patterns are compatible with agent frameworks such as LangGraph and WayFlow.

Two API design features support predictable integration. First, the API exposes both global client methods and thread-level methods. Global methods support lifecycle management and broader scoped operations. Thread-level methods reduce repetition of thread identifiers in turn-by-turn code. Second, search and memory writes accept explicit scope and exact-match controls. A caller can restrict retrieval to the current thread, broaden to the current user across threads, or widen search further when recall is prioritized over strict locality.

The workflow also encodes conservative defaults. Message and thread-history accessors are bounded rather than returning unconstrained dumps by default. Search returns a bounded number of results unless instructed otherwise. Scope-aware retrieval reduces accidental cross-user or cross-agent leakage. Schema creation is explicit policy rather than hidden behavior. These defaults reflect deployment constraints associated with scale, noise, and privacy.

%% file: sections/06_use_cases.tex
\section{Use Cases}
This section describes representative workloads that motivate the memory model and interface design of Oracle Agent Memory. Each workload emphasizes a different combination of short-term state, durable factual memory, procedural memory, and scope control.

\subsection{Conversational Continuity Within One Thread}
The first workload is continuity within a long-running interaction. A support or assistant workflow may extend across many turns while the user refines a goal, corrects assumptions, and adds constraints. Without short-term memory, the agent may lose the thread topic, repeat clarifying questions, or carry excessive raw history into the prompt, leading to higher latency, higher token cost, and greater risk of inconsistency.

In Oracle Agent Memory, this workload is handled primarily through thread summaries and context cards. Messages are added to a thread, persisted as raw evidence, and used to produce compressed summaries or structured context cards that capture thread topic, current state, and relevant facts or guidelines. The next-turn prompt is thereby reduced and stabilized, while the raw messages remain available for later retrieval.

\subsection{Personalization and Long-Term Factual Memory Across Sessions}
A second workload is personalization across repeated sessions with the same user. Users expect agents to retain preferences such as response style, domain vocabulary, recurring constraints, or durable biographical facts relevant to the interaction. If every session begins from zero, the system remains effectively stateless despite having observed sufficient information to adapt.

Durable factual memory is central in this setting. Preferences and facts can be added explicitly through the memory API or extracted automatically from thread messages. Once stored, they can be retrieved later under user and agent scope rather than only within the originating thread. The same mechanism that supports simple preference-style memories also generalizes to richer enterprise context such as preferred output formats, business definitions, known account metadata, or persistent user constraints.

Without such memory, repeated clarification, duplicated prompts, inconsistent personalization, and prompt bloat become common because stable context must be restated manually on each turn. Scoped long-term memory converts these repeated instructions into reusable state.

\subsection{Procedural Reuse and Self-Improvement}
A third workload concerns procedural memory: retention of strategies, remediation paths, or workflows that succeeded previously. This is especially relevant for agents performing research, coding, analytics, or multi-step troubleshooting. The useful memory in these settings is often not a static fact about the user, but a lesson about future action.

Self-improvement workloads also expose important safety and evaluation difficulties. Learning directly from interaction is risky because feedback may be noisy or adversarial, regressions may accumulate silently, and there is no universal mechanism for converting outcomes into reusable procedural knowledge. Oracle Agent Memory addresses only part of this problem in its current scope, but it preserves procedural memory as a first-class category. The architecture supports explicit memory insertion, extraction from thread messages, and future richer memory artifacts or graph-aware structures. This design permits retention of guidelines such as clarification rules in NL2SQL or evidence that a remediation path resolved a recurring incident.

This workload distinguishes procedural memory from personalization. Personalization captures stable user preferences; procedural memory captures what the agent has learned to do.

\subsection{Domain-Specific Structured Memory}
A fourth workload is domain-specific memory construction, illustrated by NL2SQL and related analytics agents. In this setting, the relevant memory is not limited to chat snippets. It may include table statistics, frequently used joins, schema conventions, business rules, and interpretation guidelines for recurring ambiguous requests. A query such as ``average spending'' may refer to a local business concept that is not explicit in the schema. The memory system must therefore preserve learned interpretations in addition to literal interaction content.

This workload places direct pressure on the storage model. If memory is limited to generic conversational embeddings, representation of structured task knowledge and relationships among artifacts becomes difficult. Oracle Agent Memory addresses this requirement by placing memory on a substrate that supports relational, JSON, and vector representations, with room for future graph-aware memory structures. Even when the MVP scope is limited to conversational memory and vector search, the design leaves room for richer artifacts.

\subsection{Shared Memory Across Agents and Users}
A fifth workload is shared memory under controlled scope. Many enterprise workflows involve multiple agent roles or repeated sessions per user. A user-specific preference may be useful across many threads. An agent-level guideline may generalize across many users. Some organizational knowledge may require broader sharing, but only when retrieval and authorization policies permit it.

Broad recall and safe scoping are in tension. Shared memory increases reuse and reduces cold starts, but also increases leakage risk when scope is not explicit. Oracle Agent Memory therefore treats user, agent, and thread identifiers as first-class storage and query dimensions and exposes exact-match controls at search time. This permits narrow retrieval when precision or privacy dominates and broader retrieval when recall is the primary objective.

\subsection{Shared Requirements}
Across these workloads, the same requirements recur: preservation of raw interaction, derivation of compact short-term context, storage of durable long-term memory, and retrieval of relevant evidence under explicit scope constraints. The dominant requirement differs by workload. Conversational continuity depends on summaries and context cards. Personalization depends on durable factual memory. Self-improvement depends on procedural or experiential memory. NL2SQL and related settings stress the representational model. Shared multi-agent memory stresses scoping and governance. Oracle Agent Memory is designed to address these requirements within one architectural framework.

%% file: sections/07_evaluation.tex
\section{Evaluation and Benchmarking}
Our evaluation centers on LongMemEval, a benchmark designed around timestamped multi-session chat histories in which evidence is distributed across many distractor sessions \cite{longmemeval}. This benchmark is appropriate for the present system because it approximates a realistic retrieval regime: relevant facts are sparse, temporally distributed, and often absent from the current prompt. Its task taxonomy also differentiates direct fact lookup from temporal reasoning, cross-session synthesis, knowledge updates, and abstention.

LongMemEval is not the only relevant benchmark. Section~3 described related benchmarks, including LoCoMo and MemoryAgentBench \cite{locomo,memoryagentbench}. It is central here because Agent Memory is built around persistent threads, scoped search, and retrieval of sparse prior evidence from conversation-derived memory. The evaluation reported in this section uses a newer system configuration and augments accuracy measurements with estimated-token efficiency, prompt-caching analysis, and flat-history comparisons \cite{oracleagentmemoryblog2026}.

Because the experiments in this report were run under specific model, embedding, retrieval, scoring, and prompt-construction choices, the reported numbers should be interpreted as measurements of the evaluated configuration rather than universal system limits. Published external baselines are included for context when directly reproduced numbers are unavailable. Scores across systems are directly comparable only when dataset version, model, embedding model, retrieval budget, prompt, judge, and scoring convention are held constant.

The LongMemEval run reported here used \texttt{gpt-5.5} with xhigh reasoning, \texttt{nomic-embed-v1.5} embeddings, a DB HNSW index, and top-$K=200$. The flat-history comparison used a \texttt{gpt-5.4} judge. BEAM results are reported under explicit scoring conventions described later in this section. Additional details such as exact prompts, model snapshots, and run metadata are not fully enumerated in this report, so the results should be interpreted as configuration-specific.

\subsection{LongMemEval Accuracy}
\begin{table}[t]
\centering
\small
\begin{tabularx}{\textwidth}{lX}
\toprule
Ability area & Evaluation target \\
\midrule
Single-session recall & Recover facts, preferences, or assistant-provided information from one prior session. \\
Multi-session reasoning & Combine evidence distributed across multiple prior sessions. \\
Knowledge update & Prefer corrected or more recent information when older evidence remains present. \\
Temporal reasoning & Use timestamps and explicit time references rather than semantic similarity alone. \\
Abstention & Avoid confident answers when the required evidence is absent or ambiguous. \\
\bottomrule
\end{tabularx}
\caption{LongMemEval ability areas discussed in this report. The table is qualitative; the aggregate result in Figure~\ref{fig:blog-longmemeval-categories} is reported over the full 500-question benchmark.}
\label{tab:question-distribution}
\end{table}

Table~\ref{tab:question-distribution} summarizes the ability areas discussed in the evaluation. The categories probe distinct failure modes. Information-extraction questions test whether the system can surface a directly relevant fact. Multi-session reasoning tests whether the model can combine evidence distributed across conversations. Knowledge-update questions test whether the system can prefer corrected or recent information when older evidence remains present. Temporal reasoning requires use of chronology rather than semantic similarity alone. Abstention tests whether the system avoids confident answers when evidence is absent or ambiguous.

\begin{figure*}[t]
  \centering
  \includegraphics[width=0.88\textwidth]{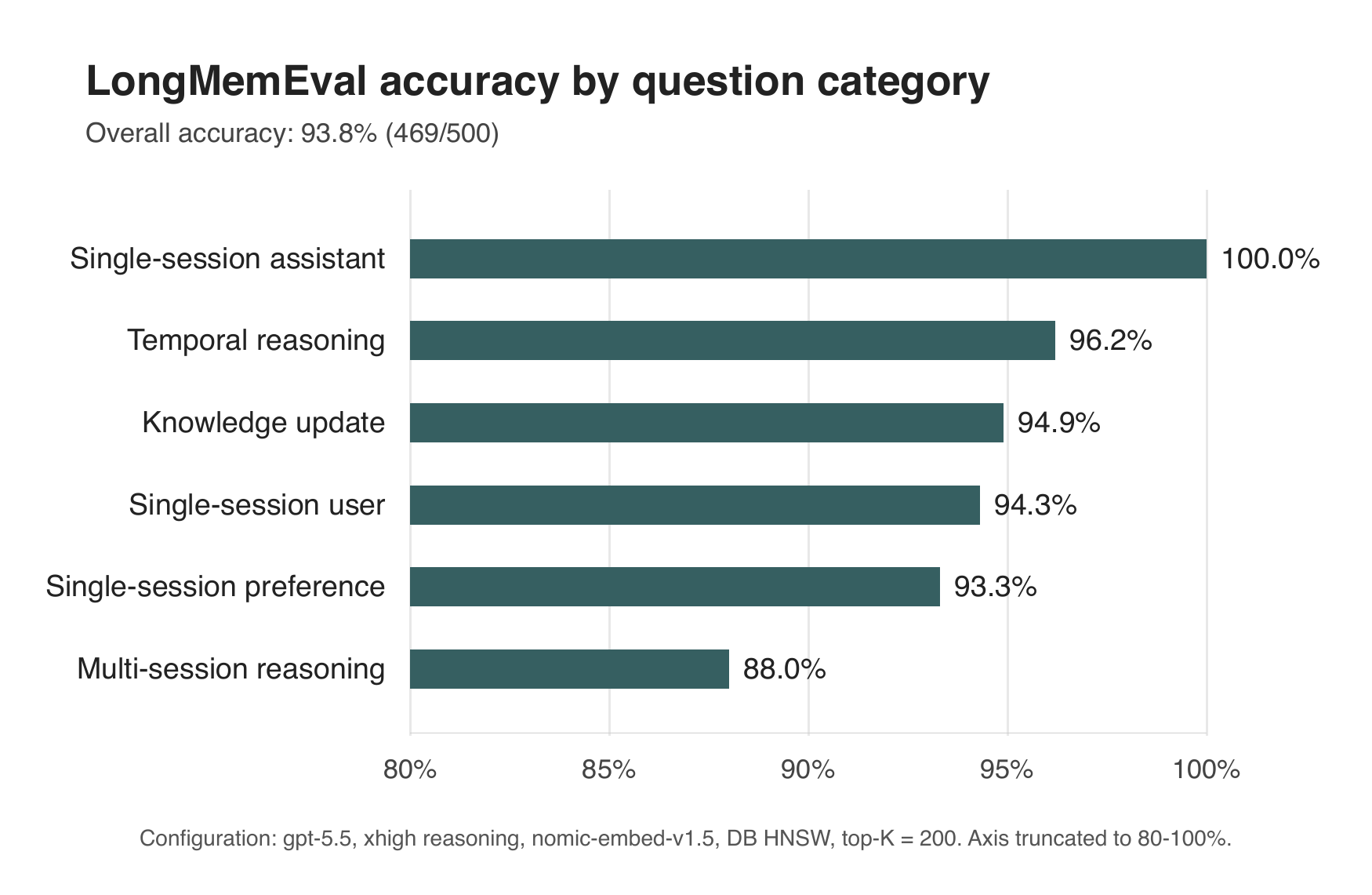}
  \caption{LongMemEval accuracy for Agent Memory under a high-accuracy evaluation configuration. Overall accuracy is 93.8\% (469/500). The run uses \texttt{gpt-5.5} with xhigh reasoning effort, \texttt{nomic-embed-v1.5} embeddings, a DB HNSW index, and top-$K=200$. The x-axis is truncated because all categories exceed 88\%.}
  \label{fig:blog-longmemeval-categories}
\end{figure*}

Figure~\ref{fig:blog-longmemeval-categories} reports the updated LongMemEval result for the high-accuracy evaluation configuration: Agent Memory reaches 93.8\% overall accuracy, corresponding to 469 correct answers out of 500. Category-level scores are 100.0\% for single-session assistant recall, 96.2\% for temporal reasoning, 94.9\% for knowledge updates, 94.3\% for single-session user recall, 93.3\% for single-session preference recall, and 88.0\% for multi-session reasoning. The multi-session category remains the lowest-scoring category, consistent with its role as the most demanding cross-session retrieval and synthesis setting. This report does not characterize the full cost-quality Pareto frontier across model choice, reasoning effort, top-$K$, reranking, and prompt construction.

\subsection{Estimated Token Behavior}
Accuracy alone does not capture the primary operational benefit of memory management. A flat-history agent appends the full verbatim transcript to each request, so per-turn input grows approximately linearly with conversation length. Agent Memory instead combines periodic thread summarization, durable memory extraction, and prompt-time context materialization to keep the working context bounded.

\begin{figure*}[t]
  \centering
  \includegraphics[width=0.9\textwidth]{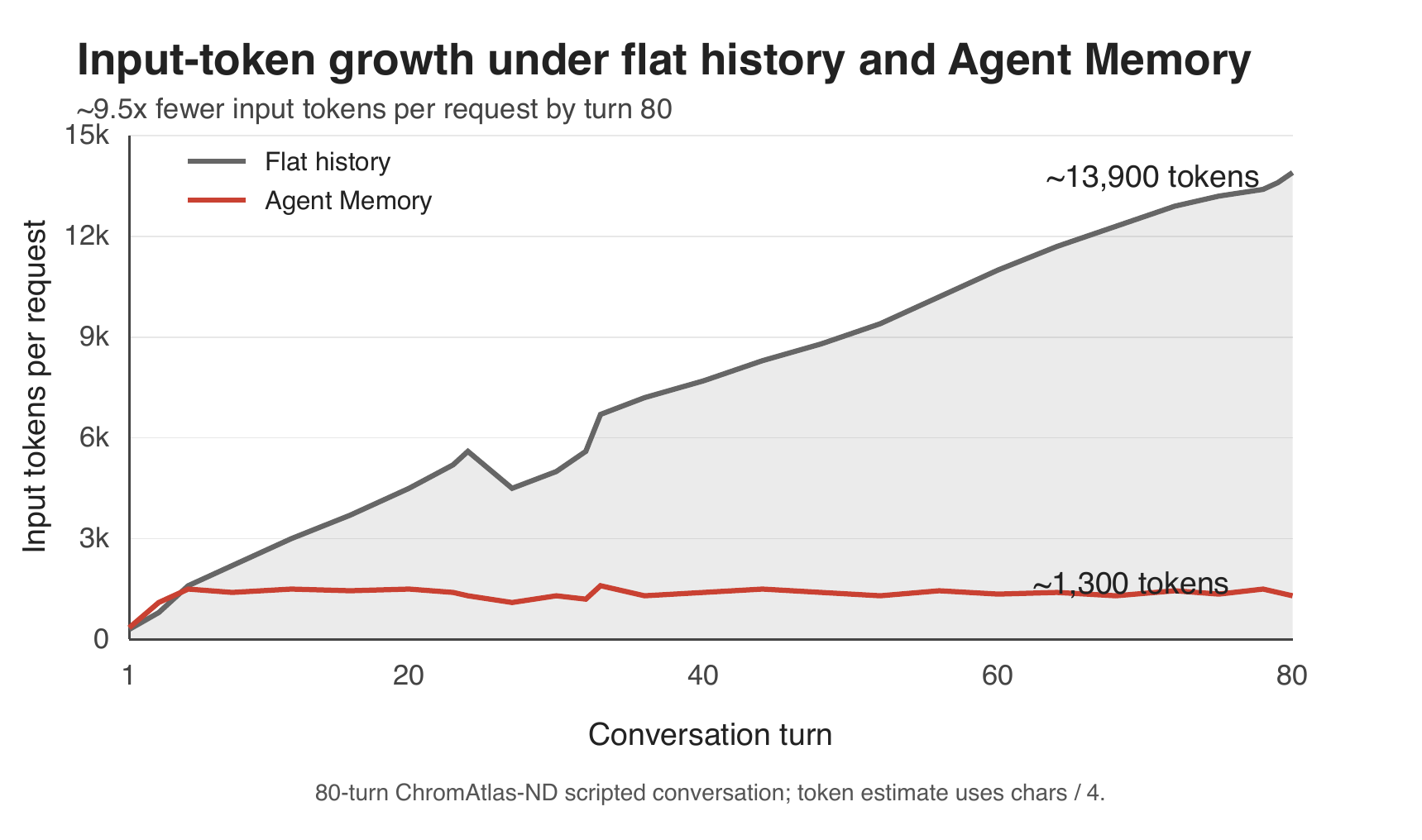}
  \caption{Input tokens per request over an 80-turn ChromAtlas-ND scripted conversation. Token estimates use the notebook convention of characters divided by four. Agent Memory remains near 1,300 input tokens per request, whereas a flat-history baseline grows to roughly 13,900 input tokens by turn 80.}
  \label{fig:blog-token-growth}
\end{figure*}

Figure~\ref{fig:blog-token-growth} shows the resulting estimated input-token behavior. By turn 80, Agent Memory holds the request near 1,300 estimated input tokens, while the flat-history baseline grows to approximately 13,900 estimated input tokens. This corresponds to roughly 10.7$\times$ fewer estimated input tokens at the final turn. These figures use approximate token estimation and should not be read as provider-billed cost measurements. The result should be interpreted as an estimated-token and context-quality effect: bounded context reduces the burden on the model to attend over a long transcript containing stale or irrelevant details.

\subsection{Comparison With Flat History}
The flat-history baseline has a strong information advantage: the full verbatim conversation is available in order. It is therefore a useful comparator for separating recall from prompt organization. A memory-managed context can lose information if summarization or retrieval fails, but it can also improve answer quality by focusing the model on the relevant state.

\begin{figure*}[t]
  \centering
  \includegraphics[width=0.78\textwidth]{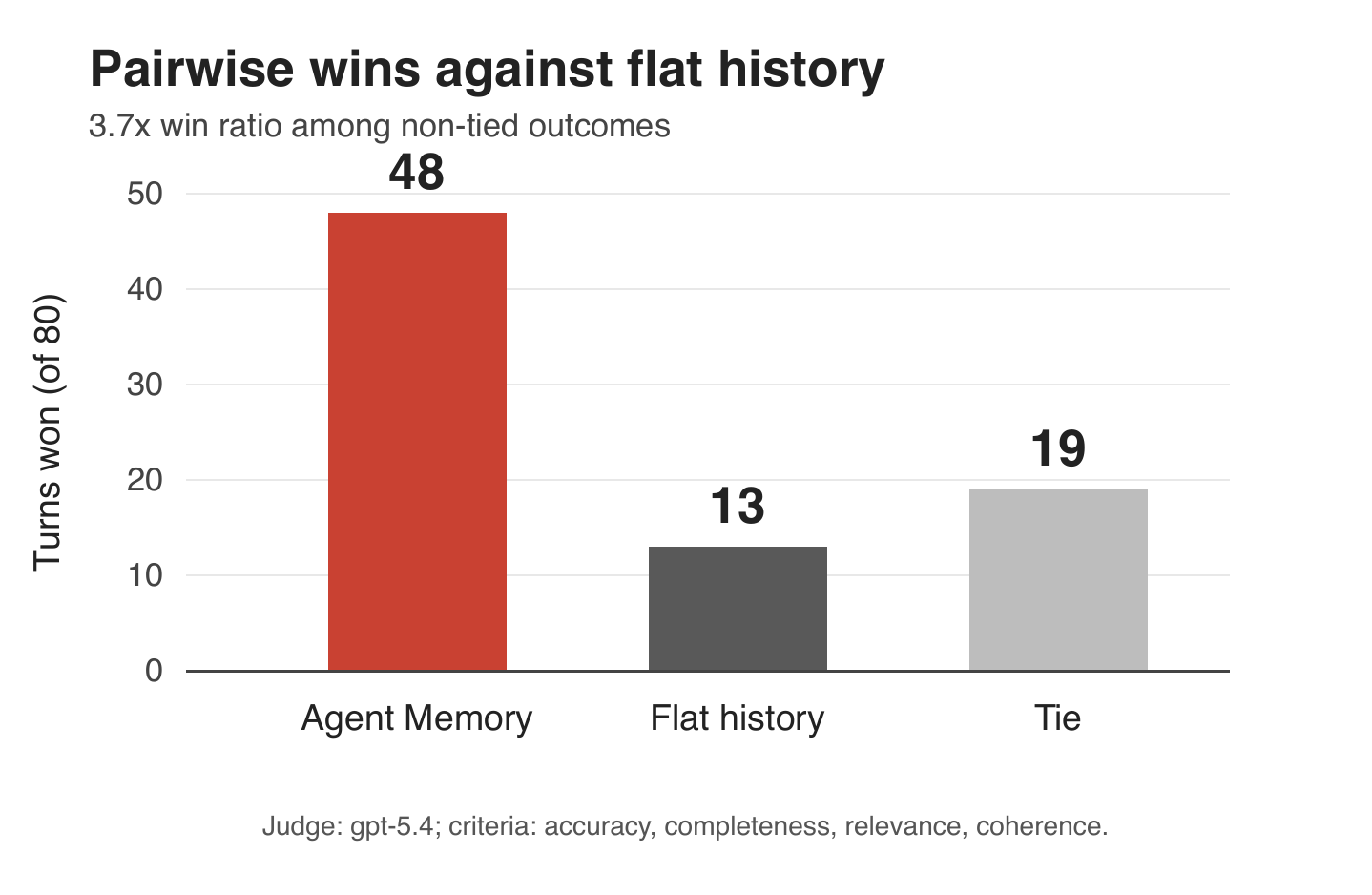}
  \caption{Pairwise judgment against a flat-history baseline over the same 80-turn ChromAtlas-ND scripted conversation. A \texttt{gpt-5.4} judge scores answers on accuracy, completeness, relevance, and coherence. Agent Memory wins 48 turns, flat history wins 13 turns, and 19 turns are ties.}
  \label{fig:blog-flat-history-wins}
\end{figure*}

Figure~\ref{fig:blog-flat-history-wins} reports the pairwise comparison. Agent Memory wins 48 of 80 turns, the flat-history baseline wins 13, and 19 are ties, for a 3.7$\times$ win ratio among non-tied outcomes. This result suggests that memory management can improve answer quality even when the baseline has access to all raw conversation content. One plausible explanation is prompt focus: a retrieved context card or bounded memory context can concentrate the model on task-relevant evidence, whereas a full transcript can distribute attention across irrelevant turns.

\subsection{Summarization Threshold Sweep}
The estimated-token and context-fidelity trade-off is tunable. Agent Memory exposes summarization and compaction behavior through thresholds that determine when thread context is compressed. Lower thresholds compact more aggressively, reducing estimated token use but increasing dependence on summaries and extracted memory. Higher thresholds preserve more raw context but reduce estimated-token benefits.

\begin{figure*}[t]
  \centering
  \includegraphics[width=0.86\textwidth]{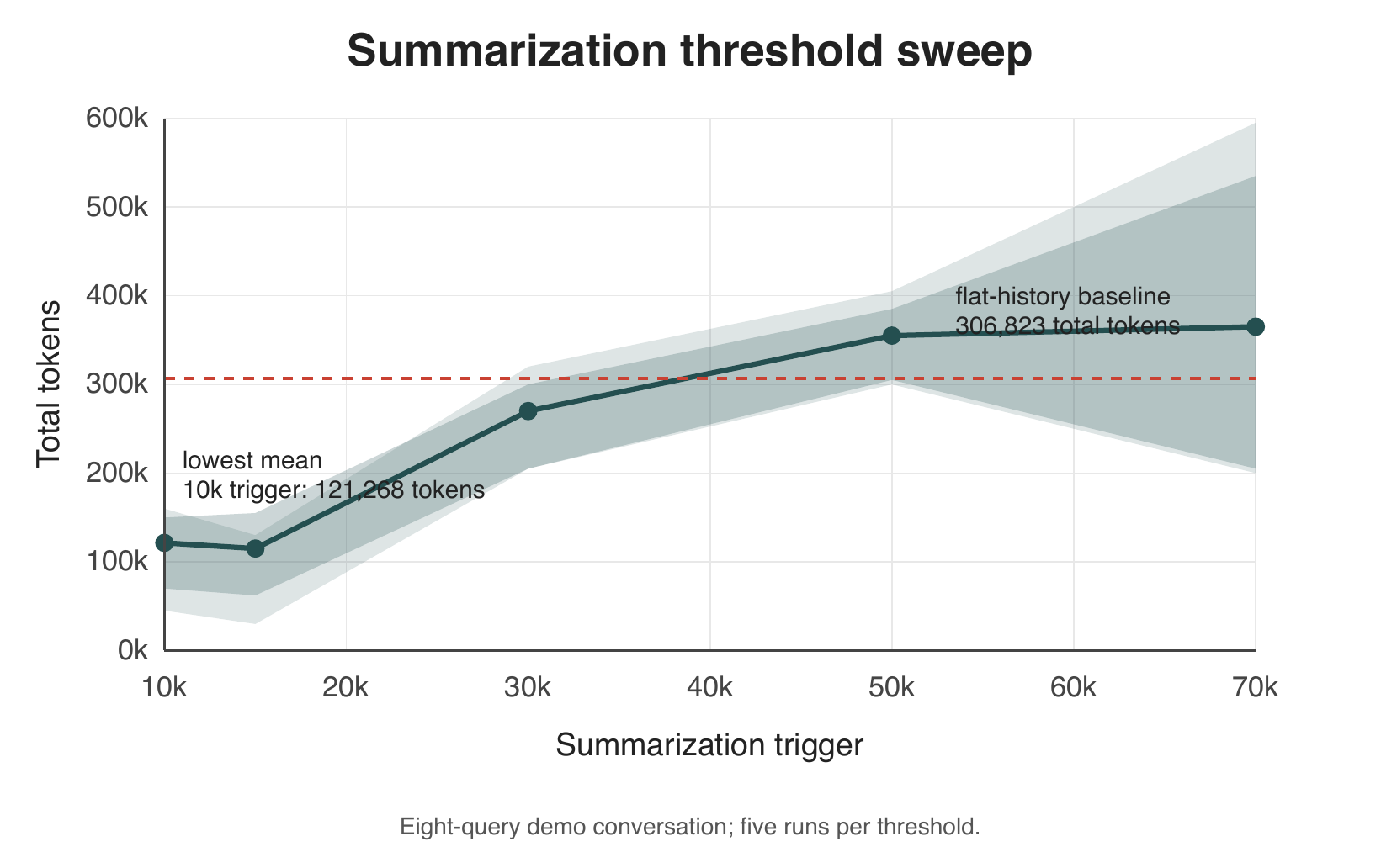}
  \caption{Summarization-trigger sweep on an eight-query demo conversation, with five runs per threshold. The flat-history baseline is 306,823 total tokens. The lowest reported mean occurs at the 10,000-token trigger, with 121,268 total tokens. Higher triggers compact less often and approach or exceed the flat-history baseline in the 50k--70k range.}
  \label{fig:blog-threshold-sweep}
\end{figure*}

Figure~\ref{fig:blog-threshold-sweep} illustrates this control surface. In the reported eight-query demo conversation, the flat-history baseline consumes an estimated 306,823 total tokens. A 10,000-token summarization trigger yields the lowest reported mean total-token estimate, 121,268 tokens. As the trigger increases, compaction occurs less often, more raw context remains in the prompt, variance widens, and mean estimated tokens approach or exceed the flat-history baseline by the 50k--70k trigger range. The threshold should therefore be viewed as an operating parameter that trades context fidelity, estimated token use, and variance.

\subsection{Prompt Caching and the Intermediate Operating Point}
Provider-side prompt caching complicates the translation from estimated token counts to effective provider cost and latency. OpenAI prompt caching is automatic for prompts of at least 1024 tokens, uses exact prefix matches, reports cache hits through \texttt{cached\_tokens}, and recommends placing static content such as instructions, tools, and examples at the beginning of the prompt while placing dynamic user-specific content later \cite{openaipromptcaching2026}. Anthropic exposes a different prompt-caching pricing model with explicit cache-write and cache-hit multipliers, indicating that caching economics are provider-specific \cite{anthropicpromptcaching2026}.

Memory compaction can reduce estimated total input tokens while simultaneously reducing cache locality. If regenerated summaries, context cards, or retrieved memory snippets appear near the beginning of the prompt, they can alter the prefix and reduce prompt-cache hit rates. In that setting, a flat-history or append-only prompt may benefit from more stable prefixes even though it consumes more estimated tokens. The evaluation objective is therefore not simply to minimize prompt length. It is to minimize effective cost and latency after accounting for estimated input tokens, cached-token discounts, cache-hit rate, and answer quality.

A practical intermediate operating point is to keep stable material at the front of the prompt and place volatile memory material later. Static instructions, tool definitions, schema, and reusable policy text should occupy the cacheable prefix. Summaries, context cards, retrieved memories, and recent user turns should appear after that prefix and should be refreshed on controlled thresholds rather than on every turn. Evaluation should track estimated input tokens, cached tokens, end-to-end latency, search latency, answer quality, and retrieval quality together. Under this policy, Agent Memory provides bounded working context while preserving provider-side prompt caching when the prompt structure and provider cache semantics allow it.

\subsection{BEAM Benchmark}
BEAM provides a complementary stress test for long-horizon conversational memory \cite{beam2026}. It contains 100 coherent multi-domain conversations at 128K, 500K, 1M, and 10M token scales, together with 2,000 human-validated probing questions. The benchmark covers ten memory abilities: abstention, contradiction resolution, event ordering, information extraction, instruction following, knowledge update, multi-session reasoning, preference following, summarization, and temporal reasoning. Compared with LongMemEval, BEAM places greater pressure on very long-context retrieval, ordering, summarization, and update handling.

The BEAM protocol also illustrates why benchmark implementation details matter. The reference evaluator floors most rubric scores from a three-point scale to binary outcomes, and one version of the judge prompt omits substitution of the original question. In addition, event-ordering scores used by one published baseline emphasize whether the correct events are mentioned, rather than whether they are returned in the correct order. We therefore report results with the scoring convention made explicit rather than treating all aggregate BEAM numbers as directly interchangeable. Event-presence and order-sensitive scores are not interchangeable. The event-presence score is included only to align with a published-style comparison; the order-sensitive score is the stricter metric for event-ordering ability.

\begingroup
\small
\begin{longtable}{p{0.09\textwidth}p{0.34\textwidth}p{0.10\textwidth}p{0.35\textwidth}}
\caption{BEAM results under explicit scoring conventions. The event-presence convention checks whether relevant events are mentioned; the order-sensitive convention additionally penalizes incorrect ordering.}
\label{tab:beam-results}\\
\toprule
Dataset & System and scoring convention & Accuracy & Notes \\
\midrule
1M & Mem0 reported baseline & 0.641 & Published-style comparison with mean answer tokens of 6,719; not independently reproduced in this run. \\
1M & Agent Memory, event-presence scoring & 0.680 & Same event-ordering convention as the published-style Mem0 comparison; 476.13/700 aggregate score. \\
1M & Agent Memory, order-sensitive scoring & 0.630 & Stricter event-ordering evaluation; 438.76/700 aggregate score. \\
10M & Agent Memory, order-sensitive scoring & 0.510 & 102.46/200 aggregate score; mean answer input tokens 122,720.8 and mean search latency 3.30s. \\
\bottomrule
\end{longtable}
\endgroup

Table~\ref{tab:beam-results} summarizes the principal measurements. Under the event-presence convention used for the published-style comparison, Agent Memory reaches 0.680 on the 1M dataset, exceeding the Mem0 reported baseline of 0.641. Under the stricter order-sensitive convention, Agent Memory reaches 0.630 on the 1M dataset. The 10M result is lower at 0.510, with category scores of 0.62 for abstention, 0.64 for contradiction resolution, 0.36 for event ordering, 0.68 for information extraction, 0.55 for instruction following, 0.65 for knowledge update, 0.17 for multi-session reasoning, 0.57 for preference following, 0.48 for summarization, and 0.40 for temporal reasoning. The 10M result should be read as an early stress-test measurement rather than a final system limit: the evaluated configuration relies heavily on retrieved windows and answer-time reasoning, so its token use remains high relative to systems that aggressively extract and consolidate memories before answering.

\begingroup
\small
\begin{longtable}{lrr}
\caption{BEAM retrieval diagnostics. Prompt recall includes evidence introduced by window expansion after the initial top-K retrieval.}
\label{tab:beam-retrieval}\\
\toprule
Question type & Top-40 recall & Prompt recall \\
\midrule
Contradiction resolution & 0.9067 & 0.9625 \\
Event ordering & 0.3920 & 0.6736 \\
Information extraction & 0.8208 & 0.9208 \\
Instruction following & 0.7375 & 0.8500 \\
Knowledge update & 0.9000 & 0.9542 \\
Multi-session reasoning & 0.6659 & 0.8590 \\
Preference following & 0.7009 & 0.8632 \\
Summarization & 0.2838 & 0.6575 \\
Temporal reasoning & 0.9250 & 0.9750 \\
\bottomrule
\end{longtable}
\endgroup

Table~\ref{tab:beam-retrieval} separates retrieval quality from answer quality. Retrieval is strong for temporal reasoning, contradiction resolution, and knowledge update, and weaker for event ordering and summarization. The gap between top-40 recall and prompt recall shows that window expansion recovers substantial additional evidence, but it also increases prompt size. Error inspection suggests several remaining bottlenecks: answer generation can fail despite sufficient evidence; knowledge updates are sometimes treated as contradictions; event-ordering tasks suffer when concrete events are compressed into broad topical summaries; and summarization questions require broader coverage than sparse keyword-oriented retrieval provides. These patterns motivate the future-work directions in Section~\ref{sec:future-work}.

These results refine the evaluation objective. LongMemEval accuracy measures whether the memory system can recover and use long-horizon evidence under sparse multi-session conditions. BEAM extends the evaluation to much longer conversations and exposes ordering, summarization, and update-resolution weaknesses. Token-growth and threshold-sweep measurements determine whether the same system remains economical in estimated-token terms as sessions extend. Flat-history comparison tests whether bounded context can improve answer quality despite receiving less raw transcript content. Prompt-caching analysis determines whether estimated-token savings translate into provider-specific latency and effective-cost benefits. A complete evaluation of agent memory must report all of these dimensions.

%% file: sections/08_discussion.tex
\section{Discussion: Why Downstream Accuracy Alone Is Not Enough}
\label{sec:discussion}
Most current agent-memory benchmarks are reported primarily through downstream accuracy on end tasks. That metric is useful, but it is not a sufficient indicator of whether a memory approach is actually working well. A correct final answer can emerge even when retrieval was noisy, over-broad, or prompt-fragile. Conversely, a wrong answer can reflect a reasoning or generation failure even when the memory subsystem surfaced the right evidence.

The present evaluation points toward memory-centric metrics such as whether the system retrieved the correct evidence. Recall-oriented diagnostics answer questions that aggregate accuracy cannot:
\begin{itemize}
  \item Did the retrieval layer surface the right prior fact, message, or summary at all?
  \item Did the system miss relevant evidence because of scope filters, top-$K$ choices, or chunking decisions?
  \item Did the agent answer correctly only because the benchmark can be solved by broad prompting rather than precise memory recovery?
  \item Did abstention behavior result from genuine absence of evidence or a memory-retrieval miss?
\end{itemize}

Additional diagnostics follow from the architecture described earlier in the report. Memory extraction quality should be evaluated separately from retrieval quality. A system may fail because the relevant fact was never promoted into durable memory, because it was chunked or indexed poorly, or because retrieval failed even though the record was present. Scope behavior should also be tested explicitly. A system that improves recall by searching too broadly may appear strong on a benchmark while remaining unsafe or unpredictable in multi-user settings. Latency and token-use measurements are equally necessary, since a system that requires high retrieval fan-out or repeated long summaries may be accurate but operationally misaligned with realistic agent loops.

Accuracy should not be discarded. Task outcome remains essential because users ultimately care about end performance. However, agent-memory evaluation should separate at least three layers: evidence retrieval (recall and precision over relevant memory), evidence use (whether the agent conditions on retrieved material correctly), and final task outcome. In more mature evaluations a fourth layer should be included as well: operational efficiency, including search latency, ingest overhead, and token cost.

Memory systems should therefore be evaluated as systems. They construct knowledge, compress context, retrieve evidence, and provide that evidence to downstream reasoning. Each of these steps may succeed or fail independently. Benchmarks that collapse them into a single correctness number remain useful, but they obscure where the bottleneck lies. For a report centered on memory architecture, that diagnostic resolution is more informative than small changes in aggregate score.

%% file: sections/09_limitations.tex
\section{Current Limitations}
\label{sec:limitations}
Several limitations bound the interpretation of the results and architectural claims in this report. First, benchmark comparisons depend on model choice, embedding model, retrieval budget, prompt construction, judge, dataset version, and scoring convention. Results from different systems are directly comparable only when these factors are held constant.

Second, this report does not fully characterize the cost-quality Pareto frontier across model choice, reasoning effort, top-$K$, reranking, summarization threshold, and prompt construction. The token measurements reported in Section~7 use approximate token estimates and should not be read as provider-billed cost measurements. Provider-side prompt caching further changes effective cost and latency in ways that depend on vendor-specific cache semantics.

Third, scope identifiers and exact-match flags are retrieval controls rather than complete authorization boundaries. Production deployments should enforce authorization through database privileges, database policy, application identity propagation, and audit controls, with memory scope filters serving as part of the retrieval contract.

Fourth, graph-aware memory evolution, deduplication, and conflict resolution remain future work or deployment-dependent capabilities. The current architecture leaves room for richer memory structures and lifecycle policies, but the present evaluation does not establish final behavior for those capabilities.

Finally, the BEAM 10M result should be read as an early stress-test measurement rather than a final system limit. It identifies useful failure modes for retrieval, ordering, summarization, and answer generation, but it does not exhaust the design space of extraction policies, memory consolidation, or lower-cost operating configurations.

%% file: sections/09_future_work.tex
\section{Future Work}
\label{sec:future-work}
The present design establishes a database-native substrate for scoped conversational memory, but several extensions are necessary for more capable long-horizon agents. The first direction is graph-aware memory. Many useful memories are relational rather than atomic: a user preference may apply only to a project, a remediation step may depend on an incident class, and an update may supersede a previous statement only under a specific temporal or task condition. Representing these dependencies as queryable relationships would permit retrieval to combine semantic similarity with graph traversal, temporal constraints, and explicit provenance over the evidence used by an agent.

A second direction is memory evolution and deduplication. Durable memory should not accumulate as an append-only log of extracted facts. It should merge repeated observations, preserve meaningful disagreements, retire stale information, and expose uncertainty when the evidence does not support a single current value. This requires update policies that distinguish contradiction from legitimate temporal change, maintain links between raw messages and derived records, and support reversible consolidation when a later interaction invalidates an earlier memory.

A third direction is more selective extraction and summarization. The BEAM results indicate that event ordering and summarization remain difficult even when relevant evidence is often retrieved. This suggests that extraction should preserve ordered event structure rather than isolated facts alone, and that summaries should be optimized for downstream question answering rather than generic compression. Future work should therefore evaluate specialized memory artifacts for timelines, procedures, preferences, and multi-step outcomes, together with retrieval policies that select among them.

A fourth direction is tighter integration between retrieval policy and security policy. In enterprise deployments, memory recall must return records that are relevant and admissible for the acting identity, agent, task, and data domain. Database-enforced policy provides the enforcement substrate, but the memory layer should make policy effects observable during evaluation: missed evidence should be distinguishable from intentionally inaccessible evidence, and retrieval diagnostics should account for both relevance and authorization.

Finally, evaluation should expand beyond aggregate answer accuracy. Future benchmarks should report memory construction quality, retrieval recall and precision, answer conditioning on retrieved evidence, estimated token use, provider-specific cache behavior, ingest overhead, and latency. LongMemEval and BEAM cover important aspects of long-horizon recall, but additional workloads are needed for multi-agent sharing, policy-constrained retrieval, graph-structured memory, adversarial or noisy feedback, and online memory evolution.

%% file: sections/09_conclusion.tex
\section{Conclusion}
Oracle Agent Memory provides a database-native substrate and API framework for scoped persistence, retrieval, summarization, and memory lifecycle management. The report argues that enterprise agent memory should be evaluated as a systems problem rather than as a prompt-construction technique alone: raw interaction must be synchronized, distilled where appropriate, retrieved under explicit scope controls, and measured by memory-centric diagnostics as well as final answer accuracy. The current design establishes the storage and integration foundation for this lifecycle while leaving richer graph-aware evolution, deduplication, conflict resolution, and broader cost-quality characterization as future work.

%% file: sections/10_appendix_setup.tex
\section{Appendix: Setup and Configuration}
Oracle Agent Memory is exposed as a Python-facing component initialized from three core dependencies: a database connection, an embedder, and an optional LLM for memory extraction. A typical setup pattern is:

\begin{lstlisting}[style=codestyle,language=Python,caption={Representative Oracle Agent Memory setup flow.}]
from oracleagentmemory.core.llms.llm import Llm
from oracleagentmemory.core.embedders.embedder import Embedder
from oracleagentmemory.core.oracleagentmemory import OracleAgentMemory
from oracleagentmemory.core import SchemaPolicy

llm = Llm(
    model="YOUR_LLM_MODEL",
    api_base="YOUR_LLM_API_BASE",
    api_key="YOUR_LLM_API_KEY",
)

embedder = Embedder(
    model="YOUR_EMBEDDING_MODEL",
    api_base="YOUR_EMBEDDING_API_BASE",
    api_key="YOUR_EMBEDDING_API_KEY",
)

memory = OracleAgentMemory(
    connection=db_pool,
    embedder=embedder,
    llm=llm,
    schema_policy=SchemaPolicy.CREATE_IF_NECESSARY,
)
\end{lstlisting}

Three implementation details are relevant in the appendix. First, the memory client is backend-agnostic at the model layer: different embedders and LLMs can satisfy the same interfaces, so the same API can support remote model endpoints or future in-database model execution paths. Second, schema creation is expressed as explicit policy rather than implicit side effect. The design supports modes that validate an existing managed schema, create objects only when the schema is empty or compatible, or recreate managed objects entirely; the default is validation-only. Third, automated memory extraction depends on whether an LLM is configured. A purely explicit memory workflow remains available in a mode that avoids automatic extraction while preserving persistence, search, and thread-level operations.

Setup also fixes several deployment assumptions. The DB user must already exist and have the privileges needed to create or validate managed tables and indexes inside that schema. The application remains responsible for database connection management and end-user access control. The memory package assumes that the caller maps end users to application-level \texttt{user\_id} values and invokes the API with the appropriate scope. These assumptions are part of the security and governance model of the overall system.

%% file: sections/11_appendix_lifecycle.tex
\section{Appendix: Memory API Lifecycle}
The API lifecycle starts from thread creation and expands outward to profiles and durable memories. The thread is the main handle through which message insertion, summarization, context-card construction, and scoped search are performed.

\begin{lstlisting}[style=codestyle,language=Python,caption={Thread lifecycle examples.}]
thread = memory.create_thread(
    thread_id="thread_123",
    user_id="user_123",
    agent_id="agent_456",
)

same_thread = memory.get_thread("thread_123")
deleted = memory.delete_thread("thread_123")

memory.add_user(
    "user_123",
    "The user prefers concise answers and works mostly with Python.",
)

memory.add_agent(
    "agent_456",
    "A coding assistant specialized in debugging and code review.",
)

memory.add_memory(
    "The user prefers short, bullet-point answers.",
    user_id="user_123",
    agent_id="agent_456",
)
\end{lstlisting}

This lifecycle separates raw conversational history from durable memory records while linking both through user, agent, and thread scope. A thread is the natural unit for turn-by-turn orchestration, but it is not the only storage scope in the system. User and agent profiles capture durable background information that may persist across many threads. Explicit memory writes allow the application or agent to persist facts, preferences, plans, or outcomes without waiting for automatic extraction from messages.

Two distinctions are relevant. First, thread lifecycle and memory lifecycle are related but not identical. Deleting a thread removes thread-scoped data, but user- or agent-level memories may remain relevant beyond that conversation depending on how they were written. Second, global client methods and thread methods expose the same underlying scope model with different interfaces. The global client is appropriate for application-level management tasks, whereas the thread handle is appropriate when the current conversation should be reused as default scope.

The lifecycle also clarifies the role of caller-provided identifiers. Threads can be created with stable thread IDs, and memories can optionally be created with custom memory identifiers. This is relevant when the application already has durable identifiers and requires idempotent or externally traceable memory records rather than opaque generated IDs. In enterprise integrations, these stable identifiers may be as important as the text payload itself.

%% file: sections/12_appendix_search.tex
\section{Appendix: Search, Context Cards, and Summaries}
The API exposes three retrieval-oriented interfaces: direct thread search, broader memory-API search with scoping options, and compact context materialization through context cards and summaries.

\begin{lstlisting}[style=codestyle,language=Python,caption={Search and short-term context patterns.}]
from oracleagentmemory.apis import Message

thread.add_messages([
    Message(role="user", content="I prefer window seats on flights."),
    {"role": "assistant", "content": "Noted. I will keep that in mind."},
])

thread.add_memory("The user is planning a trip to Kyoto.")

context_card = thread.get_context_card()
summary = thread.get_summary()

memory_results = thread.search(
    "travel preferences",
    max_results=5,
    record_types=["memory"],
)

scoped_results = memory.search(
    "What does this user prefer?",
    user_id="user_123",
    agent_id="agent_456",
    max_results=20,
)
\end{lstlisting}

These APIs instantiate the architectural claims of the main text. Working memory is materialized through summaries and context cards, whereas long-term memory is surfaced through scoped semantic retrieval over messages and durable memory records. Thread-oriented context management and database-backed retrieval are built on the same memory substrate and scope model.

Search semantics warrant particular attention. A thread-level search uses the current thread as its natural anchor, which is appropriate for turn-by-turn workflows. A broader client-level search can be expressed through explicit identifiers or through a \texttt{SearchScope} object that controls user, agent, and thread scope together with exact-match flags. Setting \texttt{exact\_thread\_match=True} yields a precision-oriented search constrained to one thread. Allowing non-exact thread matching broadens recall to other threads sharing the selected user or agent context. Record-type filters provide an additional axis of control by restricting search to messages, memories, or other supported record classes.

Context cards and summaries serve a different function. Search retrieves past evidence relevant to a query. Context cards provide compact state for the next agent turn. Summaries compress the thread so that execution can continue without replaying the full history. All three are necessary in production settings. Search is appropriate when the next step depends on a specific query. Context cards are appropriate when the agent requires a stable short-term state snapshot. Summaries are appropriate when the thread has grown long enough to require compaction under a token budget or with recent messages excluded.

The examples also illustrate several conservative defaults. Message reads are bounded by default rather than returning unconstrained history dumps. Search returns a bounded number of results unless otherwise specified. Scope is explicit and can be widened only when the caller intends it. These defaults support predictable retrieval behavior under explicit scope control.